\documentclass[letterpaper]{article} 
\usepackage{aaai2027}  
\usepackage[hyphens]{url}  
\usepackage{graphicx} 
\usepackage{natbib}  
\usepackage{caption} 
\usepackage{algorithm}
\usepackage{algorithmic}

\usepackage{amssymb}
\usepackage{amsmath}
\usepackage{multirow}
\usepackage[table,xcdraw]{xcolor}

\usepackage{newfloat}
\usepackage{listings}
\DeclareCaptionStyle{ruled}{labelfont=normalfont,labelsep=colon,strut=off} 
\floatstyle{ruled}
\newfloat{listing}{tb}{lst}{}
\floatname{listing}{Listing}

\usepackage{booktabs}

\nocopyright

\title{MedARC: Training-Free Adaptive Redundancy Compression \\of Visual Tokens for 3D Medical Vision-Language Models}
\author{
    Yitao Zhu\textsuperscript{\rm 1},
    Mengjun Liu\textsuperscript{\rm 1},
    Yingji Fu\textsuperscript{\rm 1},
    Haowen Pang\textsuperscript{\rm 2},
    Anqi Qiu\textsuperscript{\rm 1,\rm 3,\rm 4}\corresponding
}
\affiliations{
    \textsuperscript{\rm 1}Department of Health Technology and Informatics, \\The Hong Kong Polytechnic University, Hong Kong\\
    \textsuperscript{\rm 2}School of Integrated Circuits and Electronics, Beijing Institute of Technology, China\\
    \textsuperscript{\rm 3}Mental Health Research Center, \\The Hong Kong Polytechnic University, Hong Kong\\
    \textsuperscript{\rm 4}Department of Biomedical Engineering, Johns Hopkins University, USA

    an-qi.qiu@polyu.edu.hk
}

\begin{document}

\maketitle

\begin{abstract}
Integrating 3D medical images with vision-language models (VLMs) holds substantial promise for computer-aided diagnosis. However, volumetric images generate prohibitively long visual-token sequences with considerable spatial and inter-slice redundancy. Existing token compression methods typically apply uniform reduction or rely on a single importance signal, increasing the risk of removing regions that are clinically relevant to the query or structurally distinctive. To address this limitation, we propose \textbf{MedARC}, a unified, training-free framework for \textbf{A}daptive \textbf{R}edundancy \textbf{C}ompression of visual tokens in 3D medical VLMs. MedARC estimates token importance by integrating three complementary cues: self-attention from the VLM vision encoder, which reflects the model's intrinsic visual focus; similarity between projected visual tokens and text embeddings, which identifies query-relevant regions; and deviations of local visual foundation model features from the volume-level feature center, which highlight structurally distinctive anatomy. The resulting importance distribution guides a saliency-aware merging strategy that preserves informative tokens while consolidating redundant ones rather than simply discarding them. Experiments on CT-RATE and MR-RATE show that MedARC reduces visual-token overhead and inference time while preserving or improving diagnostic performance. Its multi-cue scoring cost is outweighed by the savings from processing fewer tokens, with greater benefits expected for larger language models.
\end{abstract}

\begin{links}
    \link{Code}{https://github.com/AbsterZhu/MedARC}
\end{links}

\section{Introduction}

The rapid development of multimodal large language models (MLLMs) has opened new avenues for computer-aided diagnosis, including medical report generation and visual question answering (VQA) from three-dimensional scans~\cite{zhao2024chatcad+,xin2025med3dvlm,bai2024m3d}. By connecting volumetric visual encoders with large language models (LLMs), 3D medical vision language models (VLMs) enable joint interpretation of computed tomography (CT) and clinical test, as well as magnetic resonance imaging (MRI) and clinical text. However, extending VLMs from conventional 2D images to 3D scans poses a major efficiency challenge: a single scan may contain hundreds of slices and produce thousands of visual tokens, many of which describe homogeneous tissue or highly correlated anatomical regions. Processing the full sequence increases computation, memory consumption, and context length, limiting the practical deployment of 3D medical VLMs.

Visual-token compression is therefore essential, but learning a dedicated compression module for every model or task requires additional training data, computation, and model-specific adaptation. These costs are especially restrictive in medical imaging, where expert annotations are limited and new scanners, modalities, and clinical tasks continually change the deployment setting. A training-free method can instead be applied directly to a pretrained VLM without modifying its parameters, making compression substantially easier to adopt and transfer. The central difficulty lies in preserving diagnostic evidence: information is highly nonuniformly across a volume, and subtle abnormalities may occupy only a few spatial locations or slices amid substantial redundancy~\cite{khaki2025sparsevila,liu2026medpruner}. Uniform or aggressive reduction can suppress fine-grained pathological details~\cite{yoon2025visual}. Effective training-free compression should therefore preserve input-specific diagnostic evidence while consolidating redundant context, rather than simply discarding it.

Recent studies have explored visual-token pruning, clustering, and merging to reduce the computational overhead of VLMs~\cite{bolya2022token,yang2025visionzip,chen2024image,zhu2025med}. In the medical domain, MedPruner performs hierarchical pruning at both the slice and token levels, revealing substantial redundancy in volumetric inputs~\cite{liu2026medpruner}. Nevertheless, pruning irreversibly removes or merges low-scoring tokens that may contain diagnostic context. Existing methods also tend to estimate importance from a single signal, such as feature similarity or model attention. Similarity captures redundancy but provides limited cues about clinical question, while vision-encoder attention reflects the VLM's intrinsic visual saliency and remains query-agnostic. Reliance on either signal alone can therefore preserve visually dominant content while overlooking task-relevant or structurally unusual evidence.

A vision foundation model (VFM) provides a complementary structural prior. Unlike medical image--text alignment methods that require additional supervision and training~\cite{xin2025med3dvlm,wu2025towards,chen2025large}, a frozen self-supervised VFM captures rich local and global anatomy without task-specific annotations~\cite{simeoni2025dinov3,xu2025generalizable}. Its volumetric features can reveal structurally distinctive regions overlooked by VLM attention or text similarity, complementing these cues for training-free token compression.

Motivated by these observations, we propose \textbf{MedARC}, a unified, training-free framework for \textbf{A}daptive \textbf{R}edundancy \textbf{C}ompression of visual tokens in 3D medical VLMs. MedARC employs multi-cue scoring that integrates self-attention from the VLM vision encoder, similarity between projected visual tokens and the input question, and structural saliency from a frozen 3D VFM~\cite{xu2025generalizable}. The fused distribution dynamically selects the smallest set of primary tokens covering a prescribed importance mass, thereby adapting the token budget to each volume. Rather than discarding the remaining tokens, MedARC assigns them to contextual anchors using similarity and merges their representations into a compact sequence. Although multi-cue scoring introduces additional computation, its overhead is outweighed by the substantial savings from processing fewer tokens, with increasingly pronounced efficiency gains for larger LLMs. 

The main contributions of this work are summarized as follows:
\begin{itemize}
    \item \textbf{MedARC}: We introduce a training-free framework for dynamically compressing the long visual sequences of 3D medical VLMs. MedARC adaptively retains primary tokens and merges redundant tokens without modifying the pretrained VLM.
    \item \textbf{Query-Aware Relevance}: We introduce visual--text similarity as a query-aware importance signal that measures the relevance of each visual token to the clinical question, which can significantly improve VQA performance.
    \item \textbf{Structural Saliency}: We leverage a frozen 3D VFM as an independent structural prior. By identifying regions that deviate from the dominant volumetric anatomy, its features provide complementary saliency without dense annotations or additional training.
\end{itemize}

\begin{figure*}[htb]
    \centering
    \includegraphics[width=0.95\linewidth]{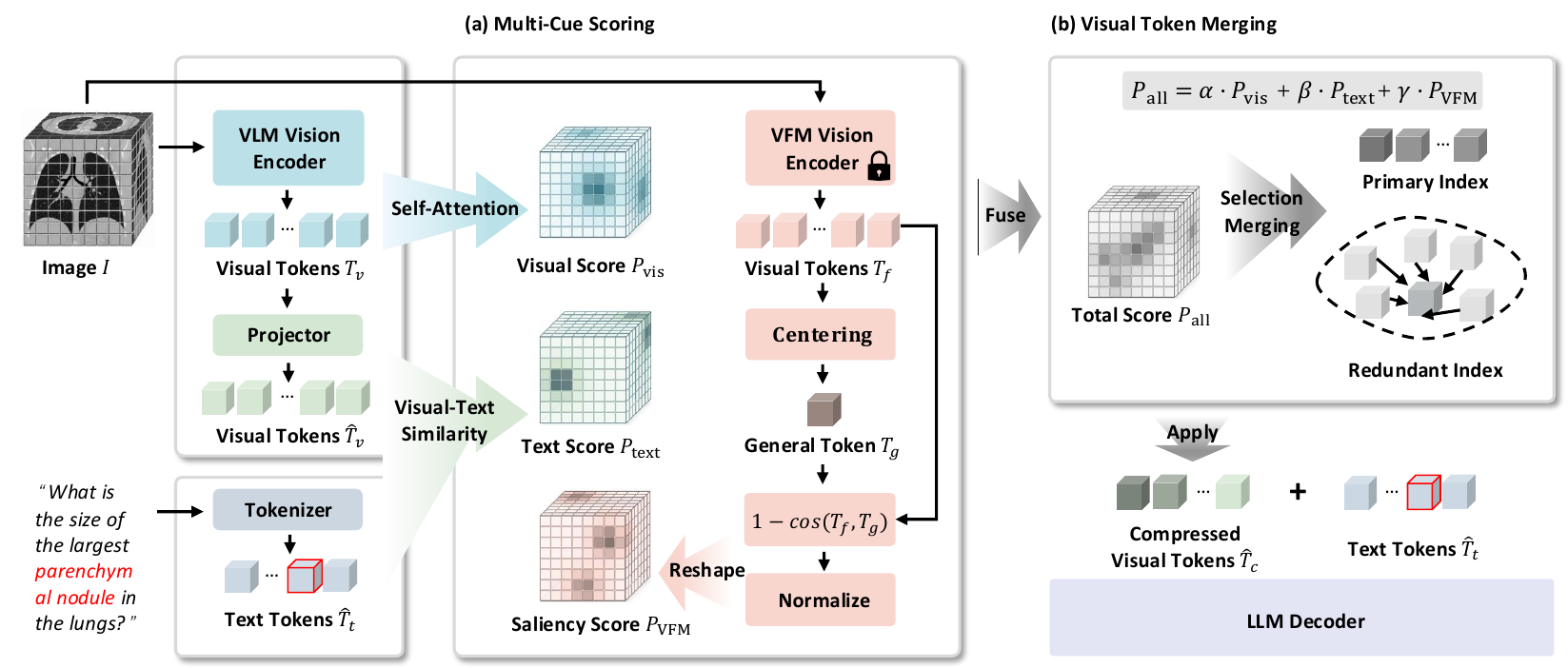}
    \caption{Overview of MedARC. (a) Visual attention, Visual--Text Similarity, and VFM Saliency provide complementary cues that are fused into a unified importance score. (b) High-importance tokens are retained based on the multi-cue score, while redundant tokens are merged before LLM decoding.}
    \label{fig:overview}
\end{figure*}

\section{Related Work}

MedARC builds on two lines of research: training-free visual-token compression and importance modeling. The former reduces redundant tokens through pruning or merging, while the latter preserves query-relevant and structurally distinctive evidence.

\subsection{Training-Free Visual-Token Compression}

Visual-token compression methods typically use learned resampling or training-free reduction. Earlier efficient vision architectures learned compact token sets or dynamically removed low-value patches, establishing adaptive tokenization and input-dependent reduction as general efficiency mechanisms~\cite{ryoo2021tokenlearner,rao2021dynamicvit}. Learned projectors map dense features to fixed-length latent tokens while preserving local context through joint multimodal optimization~\cite{cha2024honeybee,alayrac2022flamingo,li2023blip}. Although fixed-size outputs simplify batching, they require extra training, assign identical budgets to inputs of varying complexity, and may create bottlenecks under aggressive compression.

Training-free methods operate directly on pretrained VLMs. ToMe aggregates visually similar tokens to consolidate redundant content~\cite{bolya2022token}, whereas FastV prunes low-attention tokens~\cite{chen2024image}. LLaVA-PruMerge and VisionZip retain attention-dominant tokens while merging contextual ones into similar representatives~\cite{shang2025llava,yang2025visionzip}. HiPrune further uses hierarchical encoder attention to preserve object-centric anchors, adjacent buffers, and global register tokens~\cite{liu2026hiprune}. Other methods broaden the selection criteria: FitPrune calibrates layer-wise pruning to preserve attention distributions, DivPrune maximizes diversity among retained tokens, and ATP-LLaVA learns layer- and instance-specific token budgets~\cite{ye2025fit,alvar2025divprune,ye2025atp}. Nevertheless, these methods remain driven primarily by feature similarity or visual attention.

Prompt-aware methods partially address this limitation. SparseVLM uses prompt-conditioned, layer-adaptive sparsification and recycles pruned information~\cite{zhang2024sparsevlm}, showing that token importance depends on both the question and visual appearance. Earlier text-conditioned compressors similarly used linguistic context to guide patch selection or joint pruning and merging, but required task-specific optimization~\cite{jiang2022trips,cao2023pumer}. No single criterion is sufficient: similarity captures redundancy but not task relevance, visual attention reflects largely query-agnostic encoder preferences, and prompt relevance may miss structurally unusual regions with weak model attention.

This issue is critical for 3D medical scans, where many correlated slices yield long yet highly nonuniform token sequences. MedPruner reduces inter-slice and intra-slice redundancy through anchor-based filtering and cumulative-attention selection, adapting the retained budget to each volume~\cite{liu2026medpruner}. However, discarded tokens cannot contribute complementary context, while attention alone does not capture query relevance, local structural distinctiveness, or volumetric organization. MedARC therefore integrates multiple importance signals and merges redundant tokens rather than irreversibly removing them.

\subsection{Visual Importance Modeling}

Preserving subtle pathology requires selecting clinically relevant evidence, not merely detecting redundant tokens. Medical VLMs often learn global image–text alignment from scan-level descriptions~\cite{xin2025med3dvlm,wu2025towards}, but this supervision may fail to capture localized abnormalities. More precise anatomical correspondence can be learned through region-level, phrase-level, or entity-level alignment, including global and local contrastive learning between words and regions, semantic phrase grounding, and knowledge-guided matching between entities and image patches~\cite{huang2021gloria,boecking2022making,wu2023medklip,chen2025large}. However, these approaches require additional supervision and optimization, which limits their applicability to plug-and-play compression of pretrained models.

Vision foundation models provide an alternative source of fine-grained visual knowledge. VIRAL aligns intermediate MLLM representations with pretrained VFM features, demonstrating that vision-centric supervision can preserve details weakened during multimodal training~\cite{yoon2025visual}; however, it targets 2D natural images and introduces an additional training objective. Self-supervised VFMs more generally encode both local structure and global semantics without dense task-specific annotations~\cite{simeoni2025dinov3}, offering structural evidence independent of a particular VLM's attention preferences. In volumetric medical imaging, self-supervised pretraining has progressively exploited recurrent anatomy, cross-dimensional self-distillation, and subvolume position priors to learn transferable 3D representations without dense labels~\cite{zhou2019models,xie2022unimiss,wu2024voco}.

Existing work has not explored frozen 3D VFMs as structural priors for training-free token compression. MedARC combines their structural distinctiveness with VLM attention and visual--text relevance, producing a unified saliency distribution for dynamic token allocation and merging.

\begin{table*}[ht]
\centering
\scalebox{0.9}{
\begin{tabular}{lcccccccccc}
\hline
\multicolumn{1}{c}{Method} & BL-1           & BL-2           & ROUGE          & METEOR         & BERTScore      & ROUGE-1        & ROUGE-2        & BERTScore-R    & Rate$\downarrow$   & Time$\downarrow$\\ \hline
\multicolumn{10}{c}{CT-RATE Report}                                                                                                                                         \\
\rowcolor[HTML]{EFEFEF} 
Original                   & 51.55          & 41.92          & 44.45          & 44.72          & 89.37          & 58.17          & 38.06          & 88.62          & 100.00 & 11.31\\ \hline
Hulu-L1                    & 22.56          & 16.86          & 23.43          & 23.78          & 85.14          & 34.96          & 19.13          & 83.14          & 70.95  & 4.58\\
VisionZip                  & 50.40          & 40.99          & 43.99          & 44.07          & 89.34          & 57.62          & 37.57          & 88.58          & 49.31  & 6.86\\
Hiprune                    & 49.89          & 40.78          & 43.40          & 43.35          & 89.26          & 56.94          & 36.94          & 88.46          & 33.33  & 6.81\\
MedPruner                  & \textbf{51.70} & \textbf{42.04} & 44.16          & 44.42          & 89.32          & 57.65          & 37.80          & 88.59          & 78.73  & 7.43\\
\textbf{MedARC*}           & 50.51          & 41.25          & \underline{44.55}    & \underline{44.52}    & \underline{89.43}    & \textbf{58.34} & \underline{38.31}    & \underline{88.62}    & 53.02  & 6.89\\
\textbf{MedARC}            & \underline{51.08}    & \underline{41.72}    & \textbf{44.75} & \textbf{44.72} & \textbf{89.44} & \underline{58.28}    & \textbf{38.37} & \textbf{88.65} & 79.66  & 7.18\\
 \hline
\multicolumn{10}{c}{MR-RATE Report}                                                                                                                                         \\
\rowcolor[HTML]{EFEFEF} 
Original                   & 49.58          & 40.86          & 45.48          & 45.13          & 89.52          & 55.42          & 36.66          & 88.47          & 100.00 & 11.56\\ \hline
Hulu-L1                    & 33.58          & 24.59          & 32.49          & 33.33          & 87.47          & 44.11          & 22.61          & 86.13          & 66.91  & 4.63\\
VisionZip                  & 48.33          & 39.94          & 45.51          & 44.64          & \textbf{89.54}          & 55.05          & 36.54          & 88.39          & 49.09  & 7.51\\
Hiprune                    & 46.70          & 38.46          & 44.93          & 43.80          & 89.44          & 54.46          & \textbf{38.88}          & 88.23          & 33.64  & 7.20\\
MedPruner                  & \underline{49.61}          & \underline{40.92}          & \textbf{45.58}          & \underline{45.06}          & \underline{89.53}          & \underline{55.35}          & \underline{36.70}          & \underline{88.46}          & 86.86  & 7.70\\
\textbf{MedARC*} & 48.89 &40.31 &45.20 &44.71 &89.49 &55.08 &36.32 &88.44 & 55.93 & 7.45\\
\textbf{MedARC}            & \textbf{49.82}          & \textbf{41.06}          & \underline{45.44}          & \textbf{45.17}          & \textbf{89.54}          & \textbf{55.45}          & \underline{36.70}          & \textbf{88.49}          & 80.50  & 7.71\\
\hline
\multicolumn{1}{c}{Method} & BL-1           & BL-2           & ROUGE          & METEOR         & BERTScore      & Accuracy            & F1-Score             & MAE            & Rate$\downarrow$   & Time $\downarrow$\\ \hline
\multicolumn{10}{c}{CT-RATE VQA}                                                                                                                                            \\
\rowcolor[HTML]{EFEFEF} 
Original                   & 51.93          & 42.18          & 58.61          & 58.56          & 93.49          & 70.29          & 63.78          & 6.99         & 100    & 1.39\\ \hline
Hulu-L1                    & 52.50          & 41.84          & 57.89          & 58.00          & 93.38          & \textbf{70.60} & 64.01          & 7.04           & 70.76  & 1.29\\
VisionZip                  & 51.51          & 41.67          & \textbf{58.54} & 58.28          & 93.46          & 70.08          & 63.21          & 8.22           & 49.31  & 1.28\\
Hiprune                    & 50.17          & 40.55          & 58.12          & 57.73          & 93.40          & 69.75          & 63.23          & 7.91           & 33.33  & 1.25\\
MedPruner                  & \underline{52.01}    & \textbf{42.24} & \underline{58.45}    & \underline{58.36}    & \underline{93.47}    & 70.06          & 63.21          & 7.08           & 78.71  & 0.79\\
\textbf{MedARC*}           & \textbf{52.12} & 42.01          & 58.25          & 58.22          & 93.44          & 70.49          & \underline{64.12}    & \underline{6.98}     & 52.66  & 1.35\\ 
\textbf{MedARC}            & \underline{52.01}    & \underline{42.19}    & 58.44          & \textbf{58.38} & \textbf{93.48} & \underline{70.52}    & \textbf{64.40} & \textbf{6.88}  & 79.32  & 0.90\\
\hline
\end{tabular}
}
\caption{Results on the CT-RATE and MR-RATE validation sets. \textbf{Rate} denotes the visual-token retention rate, and \textbf{Time} denotes the average inference time in seconds with 5 iterations. \textbf{Bold} and \underline{underlined} values indicate the best and second-best results, respectively. The asterisk (*) denotes an aggressive MedARC setting using a lower cumulative-importance threshold \(\tau\).}
\label{tab:comparison}
\end{table*}

\section{Method}
\label{sec:method}


MedARC is a training-free visual-token compression framework designed for volumetric medical vision--language models. Given a medical volume \(I\) and a text query \(q\), the frozen VLM vision encoder first converts the volume into visual tokens
\(T_v=\{t_{v,n}\}_{n=1}^{N}\). A multimodal projector then maps these tokens into the LLM embedding space, producing
\(\widehat{T}_v=\{\widehat{t}_{v,n}\}_{n=1}^{N}\).
Meanwhile, the tokenizer converts the query into text embeddings
\(\widehat{T}_t=\{\widehat{t}_{t,m}\}_{m=1}^{M}\).


As shown in Figure~\ref{fig:overview}, MedARC has two stages. In panel (a), visual self-attention, visual--text similarity, and 3D VFM saliency produce \(P_{\mathrm{vis}}\), \(P_{\mathrm{text}}\), and \(P_{\mathrm{VFM}}\), which are fused into a unified importance score \(P_{\mathrm{all}}\). In panel (b), high-scoring tokens are retained as primary tokens, while the remaining tokens are consolidated through similarity-based merging. The compressed sequence \(\widehat{T}_c\) is then concatenated with the text tokens and passed to the LLM decoder. We detail each component in the following sections.

\subsection{Multi-Cue Token Scoring}
\label{sec:importance_estimation}

The three scoring branches in Figure~\ref{fig:overview}(a) are computed independently and then aligned to the same visual-token grid. 

\paragraph{Visual Score.}
The first branch measures the importance of each image region according to the VLM's own visual attention inspired by MedPruner~\cite{liu2026medpruner}. We extract the self-attention matrices from the final vision-encoder block. For attention head \(h\), the attention matrix is

\begin{equation}
A_h=
\operatorname{softmax}_{\mathrm{key}}
\left(
\frac{Q_hK_h^\top}{\sqrt{d_h}}
\right),
\label{eq:visual_attention}
\end{equation}
where \(Q_h\), \(K_h\), and \(d_h\) denote the query features, key features, and feature dimension of head \(h\), respectively.

A visual token is considered important when it receives strong attention from other visual tokens. We therefore sum the incoming attention of each token and average it across all \(H_a\) attention heads:

\begin{equation}
s_n^{\mathrm{vis}}
=
\frac{1}{H_a}
\sum_{h=1}^{H_a}
\sum_i A_{h,i,n}.
\label{eq:visual_attention_score}
\end{equation}

The attention scores are reshaped and averaged according to the token organization of the multimodal projector. We denote this alignment operation by \(\mathcal{R}_v(\cdot)\). The normalized visual score is then

\begin{equation}
P_{\mathrm{vis}}
=
\operatorname{softmax}
\left(
\frac{\mathcal{R}_v(
\{s_n^{\mathrm{vis}}\})}
{T_{\mathrm{vis}}}
\right),
\label{eq:visual_score}
\end{equation}
where \(T_{\mathrm{vis}}\) is a temperature parameter. A smaller temperature produces a more concentrated distribution, whereas a larger temperature distributes the visual importance more evenly.

This branch reflects the intrinsic preference of the target vision encoder. It is useful for identifying visually prominent regions, but it does not explicitly consider the user's question nor overcomes the limitations of the VLM’s relatively coarse visual representations.

\paragraph{Text Score.}
To make the token selection dependent on the query, the second branch compares the projected visual tokens \(\widehat{T}_v\) with the text tokens \(\widehat{T}_t\). Since both sets of tokens lie in the LLM embedding space, their similarity can be computed directly.

For each visual token, we retain its highest cosine similarity to any text token:

\begin{equation}
s_n^{\mathrm{text}}
=
\max_{1\leq m\leq M}
\operatorname{cos}
\left(
\widehat{t}_{v,n},
\widehat{t}_{t,m}
\right).
\label{eq:text_similarity}
\end{equation}

The text score is obtained by normalizing these similarities over all visual tokens:

\begin{equation}
P_{\mathrm{text}}
=
\operatorname{softmax}
\left(
\frac{
\{s_n^{\mathrm{text}}\}_{n=1}^{N}
}{
T_{\mathrm{text}}
}
\right).
\label{eq:text_score}
\end{equation}

Using the maximum similarity allows a single clinically meaningful word or phrase, such as an anatomical location, lesion type, or measurement-related term, to highlight its corresponding image region. For example, when the query asks about a pulmonary nodule, tokens associated with the lung and nodule regions can receive higher scores even when they are not the most visually dominant regions in the entire volume. This query-aware emphasis can substantially improve performance on question-driven tasks such as VQA.

\begin{figure*}[!th]
    \centering
    \includegraphics[width=1\linewidth]{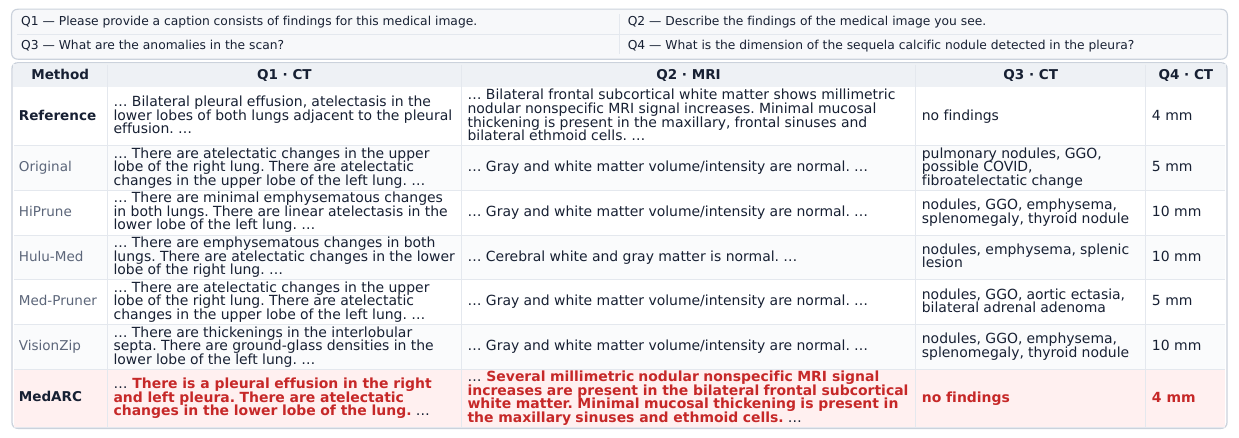}
    \caption{Qualitative comparison of MedARC with the uncompressed model
    (\emph{Original}) and training-free token-compression baselines on CT-RATE
    and MR-RATE. Q1 and Q2 show CT and MRI report-generation examples,
    while Q3 and Q4 present CT-RATE VQA cases. MedARC outputs are highlighted.}
    \label{fig:example}
\end{figure*}

\paragraph{VFM Saliency Score.}
The third branch uses a frozen 3D visual foundation model (3DINO~\cite{xu2025generalizable}) to identify locally distinctive anatomical content. The VFM extracts a set of volumetric features $T_f=\{t_{f,k}\}_{k=1}^{N_f}$ from the same image $I$. Following the centering operation shown in Figure~\ref{fig:overview}(a), we first normalize the local features and compute a general token \(T_g\) representing the dominant feature direction of the entire volume:

\begin{equation}
\bar{t}_{f,k}
=
\frac{t_{f,k}}
{\|t_{f,k}\|_2},
T_g
=
\frac{
\sum_{k=1}^{N_f}\bar{t}_{f,k}
}{
\left\|
\sum_{k=1}^{N_f}\bar{t}_{f,k}
\right\|_2
}.
\label{eq:general_token}
\end{equation}

We then measure how different each local VFM token is from this general token:

\begin{equation}
s_k^{\mathrm{VFM}}
=
1-
\operatorname{cos}
\left(
\bar{t}_{f,k},
T_g
\right).
\label{eq:vfm_saliency}
\end{equation}

A high value indicates that the local feature differs from the dominant anatomical pattern of the volume. Such regions are less likely to be repetitive background and are therefore assigned greater structural importance.

Because the VFM and VLM may produce different 3D feature grids, the scalar saliency map is reshaped and interpolated to the resolution of \(\widehat{T}_v\). We denote this operation by \(\mathcal{R}_f(\cdot)\). The normalized VFM score is

\begin{equation}
P_{\mathrm{VFM}}
=
\operatorname{softmax}
\left(
\frac{
\mathcal{R}_f(
\{s_k^{\mathrm{VFM}}\}_{k=1}^{N_f}
)
}{
T_{\mathrm{VFM}}
}
\right).
\label{eq:vfm_score}
\end{equation}

Unlike \(P_{\mathrm{vis}}\) and \(P_{\mathrm{text}}\), this score is independent of the attention behavior and multimodal alignment of the target VLM. It therefore provides an additional structural signal for protecting locally unusual or informative regions.

\paragraph{Score Fusion.}
After reshaping, the three score distributions are defined on the same \(N\)-token grid. We combine them using the weighted sum shown in Figure~\ref{fig:overview}:

\begin{equation}
P_{\mathrm{all}}
=
\alpha P_{\mathrm{vis}}
+
\beta P_{\mathrm{text}}
+
\gamma P_{\mathrm{VFM}},
\label{eq:score_fusion}
\end{equation}
where $\alpha,\beta,\gamma\geq 0$, $\alpha+\beta+\gamma=1$.

Here, \(P_{\mathrm{vis}}\) captures the VLM's intrinsic visual preference, \(P_{\mathrm{text}}\) captures query-related evidence, and \(P_{\mathrm{VFM}}\) captures structurally distinctive content. Since all three scores are normalized before fusion, the coefficients directly control their relative contributions. 

\begin{table*}[!thp]
\centering
\scalebox{0.9}{
\begin{tabular}{lcccccccccc}
\hline
\multicolumn{1}{c}{Method} & BL-1  & BL-2  & ROUGE & METEOR & BERTScore & ROUGE-1 & ROUGE-2 & BERTScore-R & Rate$\downarrow$   & Time$\downarrow$\\ \hline
\multicolumn{10}{c}{CT-RATE Report}                                                                                \\
\rowcolor[HTML]{EFEFEF}
Original                   & 51.55 & 41.92 & 44.45 & 44.72 & 89.37 & 58.17 & 38.06 & 88.62 & 100.00 & 11.31\\ \hline
Visual Att                 & \textbf{51.63} & \textbf{41.97} & 44.27 & \underline{44.50} & 89.34 & 57.77 & 37.88 & \underline{88.59} & 78.57 & 7.06\\
+ Text Sim                 & 50.73 & 41.41 & \underline{44.48} & 44.38 & \underline{89.38} & \underline{58.00} & \underline{38.15} & 88.57 & 78.05 & 7.19\\
+ VFM                      & \underline{51.08} & \underline{41.72} & \textbf{44.75} & \textbf{44.72} & \textbf{89.44} & \textbf{58.28} & \textbf{38.37} & \textbf{88.65} & 79.66 & 7.18\\ \hline
\multicolumn{10}{c}{MR-RATE Report}                                                                                \\
\rowcolor[HTML]{EFEFEF}
Original                   & 49.58 & 40.86 & 45.48 & 45.13 & 89.52 & 55.42 & 36.66 & 88.47 & 100.00 & 11.56\\ \hline
Visual Att                 & \underline{49.65} & \underline{40.94} & \underline{45.49} & \underline{45.10} & 89.52 & 55.37 & 36.65 & \underline{88.47} & 86.90 & 7.82\\
+ Text Sim                 & 49.12 & 40.6 & \textbf{45.65} & 45.08 & \textbf{89.55} & \underline{55.39} & \textbf{36.77} & 88.46 & 78.80 & 7.67\\
+ VFM                      & \textbf{49.82} & \textbf{41.06} & 45.44 & \textbf{45.17} & \underline{89.54} & \textbf{55.45} & \underline{36.70} & \textbf{88.49} & 80.50 & 7.71\\ \hline
\multicolumn{1}{c}{Method} & BL-1  & BL-2  & ROUGE & METEOR & BERTScore & ACC & F1 & MAE & Rate$\downarrow$ & Time$\downarrow$\\ \hline
\multicolumn{10}{c}{CT-RATE VQA}                                                                                   \\
\rowcolor[HTML]{EFEFEF}
Original                   & 51.93 & 42.18 & 58.61 & 58.56 & 93.49 & 70.29 & 63.78 & 6.99 & 100 & 1.39\\ \hline
Visual Att                 & \textbf{52.07} & \underline{42.18} & 58.36 & 58.3 & \underline{93.45} & 69.96 & 63.69 & \underline{6.94} & 78.55 & 0.91\\
+ Text Sim                 & 51.56 & 41.84 & \textbf{58.66} & \textbf{58.46} & \textbf{93.48} & \underline{70.47} & \textbf{64.59} & 7.64 & 77.89 & 0.85\\
+ VFM                      & \underline{52.01} & \textbf{42.19} & \underline{58.44} & \underline{58.38} & \textbf{93.48} & \textbf{70.52} & \underline{64.40} & \textbf{6.88} & 79.32 & 0.90\\ \hline
\end{tabular}
}
\caption{Ablation of the three importance signals at their default operating points. Visual Att uses vision-encoder attention, + Text Sim adds visual--text relevance, and + VFM further adds structural saliency from the frozen 3D VFM. \textbf{Bold} and \underline{underlined} values denote the best and second-best results among the compressed variants, respectively.}
\label{tab:ablation}
\end{table*}

\subsection{Adaptive Token Selection and Merging}
\label{sec:token_compression}

As shown in Figure~\ref{fig:overview}(b), the total score \(P_{\mathrm{all}}\) partitions visual tokens into primary and redundant sets. Inspired by VisionZip~\cite{yang2025visionzip}, MedARC extends adaptive token allocation from 2D images to 3D medical volumes, determining the token budget from the importance distribution of each input rather than using a fixed compression ratio.

Let \(\sigma\) denote the indices of \(P_{\mathrm{all}}\) sorted in descending order. Given a cumulative importance threshold \(\tau\), the number of primary tokens is

\begin{equation}
K
=
\min
\left\{
k:
\sum_{r=1}^{k}
P_{\mathrm{all},\sigma(r)}
\geq \tau
\right\}.
\label{eq:number_primary}
\end{equation}

Then we can get the primary index sets $\mathcal{I}{\mathrm{pri}}$ and redundant index sets $\mathcal{I}{\mathrm{red}}$, 
this rule retains fewer tokens when importance is concentrated and more when relevant evidence is distributed across multiple slices or anatomical regions. The primary tokens are preserved unchanged, while the redundant tokens are consolidated to retain complementary anatomical context. For the merging process, we can select $C$ contextual anchors in $\mathcal{I}{\mathrm{red}}$. 

\begin{equation}
C
=
\min
\left\{
\max(\lfloor\rho N\rfloor,1),
|\mathcal{I}_{\mathrm{red}}|
\right\},
\label{eq:number_context}
\end{equation}

where \(\rho\) controls the contextual-token budget, with \(C=0\) when \(\mathcal{I}_{\mathrm{red}}\) is empty. The anchor indices
\(\mathcal{I}_{\mathrm{ctx}}\subseteq\mathcal{I}_{\mathrm{red}}\)
are sampled approximately uniformly in the original volumetric order to provide broad spatial and slice-wise coverage.

Each remaining redundant token is assigned to its most similar contextual anchor using the original vision-encoder features \(T_v\):

\begin{equation}
g(i)
=
\arg\max_{c\in\mathcal{I}_{\mathrm{ctx}}}
\operatorname{cos}(t_{v,i},t_{v,c}).
\label{eq:anchor_assignment}
\end{equation}

For each anchor \(c\), its assigned source set is

\begin{equation}
\mathcal{S}_c
=
\left\{
i\in\mathcal{I}_{\mathrm{red}}\setminus\mathcal{I}_{\mathrm{ctx}}
:
g(i)=c
\right\}.
\label{eq:source_set}
\end{equation}

The contextual representation is then updated as

\begin{equation}
\widehat{t}_{c,c}
=
\begin{cases}
\displaystyle
\widehat{t}_{v,c}
+
\frac{1}{|\mathcal{S}_c|}
\sum_{i\in\mathcal{S}_c}\widehat{t}_{v,i},
&
|\mathcal{S}_c|>0,\\[6pt]
\widehat{t}_{v,c},
&
|\mathcal{S}_c|=0.
\end{cases}
\label{eq:redundant_merging}
\end{equation}

This similarity-based assignment allows visually related tokens to be merged regardless of their separation in the flattened sequence. The residual formulation in Equation~\eqref{eq:redundant_merging} preserves the original anchor representation while integrating complementary information from its assigned tokens. Finally, the primary tokens and context-enhanced anchors are restored to their original volumetric order, yielding the compressed token sequence denoted by \(\widehat{T}_c\in\mathbb{R}^{K+C}\).


The resulting sequence contains \(K+C\) tokens instead of \(N\). As shown at the bottom of Figure~\ref{fig:overview}(b), \(\widehat{T}_c\) is combined with the unchanged text sequence \(\widehat{T}_t\) and passed to the LLM decoder, preserving clinically important regions individually while summarizing repetitive anatomy.

\subsection{Implementation Details}

We implement MedARC in PyTorch using the Hugging Face Transformers library, with Qwen3-VL-4B-Instruct~\cite{bai2025qwen3} as the target VLM. Using the Qwen-VL Series Fine-Tuning framework~\cite{Qwen2-VL-Finetuning}, we independently fine-tune the model on the CT-RATE~\cite{hamamci2024foundation} and MR-RATE datasets with LoRA~\cite{hu2022lora} for five epochs, using a learning rate of \(1\times10^{-4}\).

Both the target VLM and the 3DINO~\cite{xu2025generalizable} remain frozen in compression experiments. MedARC only accesses their intermediate features to estimate token importance and merge redundant tokens at inference time. To balance importance scores, we set \(\alpha=0.4\), \(\beta=0.3\), and \(\gamma=0.3\). For adaptive compression, we fix $\rho=0.05$ and vary $\tau$ from $0.6$ to $0.9$ to obtain different compression levels.
The same weights are used in all experiments. All experiments are conducted on four NVIDIA A100 GPUs using BF16 mixed-precision computation and DeepSpeed ZeRO-2.

\section{Experiments}

\subsection{Datasets \& Evaluation Metrics}

We evaluate MedARC on CT-RATE and MR-RATE. CT-RATE~\cite{hamamci2024foundation} contains 25,692 non-contrast chest CT scans, expanded to 50,188 volumes through multiple reconstructions, from 21,304 patients. MR-RATE contains 705,254 contrast and non-contrast brain and spine MRI volumes from 98,334 studies and 83,425 patients. Both datasets provide paired radiology reports and metadata,
while CT-RATE additionally includes multi-abnormality labels. We resample all volumes to \(128\times384\times384\) (DHW).

We evaluate report generation on both datasets and VQA on CT-RATE. Generated text is assessed using BLEU-1, BLEU-2, ROUGE, METEOR, and BERTScore; task-level performance is measured using accuracy, F1-score, and MAE. We also report the visual-token retention rate to quantify compression efficiency.

\begin{figure*}[!htp]
    \centering
    \includegraphics[width=0.84\linewidth]{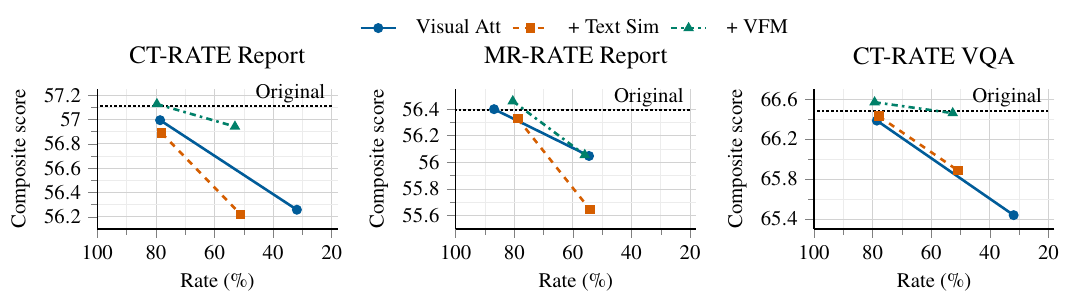}
    \caption{Compression--quality trade-offs of three importance-signal variants. Report scores average the eight metrics, while VQA scores average seven higher-is-better metrics and \(100-\mathrm{MAE}\). Curves connect the default and aggressive settings; the dotted line indicates the uncompressed model. Lower retention rates indicate fewer visual tokens.
    }
    \label{fig:ablation}
\end{figure*}

\subsection{Comparison with Other Methods}

Table~\ref{tab:comparison} compares MedARC with the uncompressed model (\emph{Original}) and four training-free baselines: Hulu-L1~\cite{jiang2025hulu}, VisionZip~\cite{yang2025visionzip}, HiPrune~\cite{liu2026hiprune}, and MedPruner~\cite{liu2026medpruner}. These methods cover similarity-based, attention-guided, and medical-specific reduction strategies. We report MedARC under different \(\tau\) settings, corresponding to token retention rates of approximately \(80\%\) and \(50\%\).

\paragraph{Report-Generation Performance.}
MedARC achieves a favorable balance between token reduction, report quality, and inference efficiency. On CT-RATE, it retains \(79.66\%\) of the tokens and achieves the best performance among the compressed methods on five of the eight metrics: ROUGE, METEOR, BERTScore, ROUGE-2, and BERTScore-R. It also outperforms the uncompressed model on five metrics while matching its METEOR score. On MR-RATE, MedARC retains \(80.50\%\) of the tokens, matches or surpasses the uncompressed model on seven metrics, and delivers the strongest overall performance. MedARC also substantially reduces generation time, indicating that the additional computation introduced by multi-cue scoring does not offset the efficiency gains. These gains are expected to become more pronounced with larger LLMs, for which visual-token processing accounts for a greater computational cost. Overall, the results suggest that primary-token retention preserves report-relevant evidence, while contextual merging recovers useful information from redundant tokens.

\paragraph{VQA Performance.}
The CT-RATE VQA results show that MedARC preserves task-level clinical performance under compression. At \(79.32\%\) retention, it improves accuracy from \(70.29\) to \(70.52\), increases the F1-score from \(63.78\) to \(64.40\), and reduces MAE from \(6.99\) to \(6.88\) relative to the uncompressed model. Even at \(52.66\%\) retention, MedARC* achieves an F1-score of \(64.12\) and an MAE of \(6.98\). Together with the report-generation results, these findings support the effectiveness of the proposed multi-cue importance scoring and retain-and-merge design.

\paragraph{Performance under Aggressive Compression.}
At comparable high-compression settings, MedARC* is most directly compared with VisionZip. On CT-RATE report generation, MedARC* retains \(53.02\%\) of the tokens and outperforms VisionZip at \(49.31\%\) retention on all eight metrics. On CT-RATE VQA, it also achieves higher BLEU-1, BLEU-2, accuracy, and F1-score, while reducing MAE from \(8.22\) to \(6.98\). Despite using substantially fewer tokens, MedARC* further surpasses MedPruner on six CT-RATE report metrics and all three principal VQA metrics. On MR-RATE, however, MedARC* at \(55.93\%\) retention underperforms VisionZip, indicating that MRI may require a larger token budget. Overall, MedARC offers a strong compression--quality trade-off on CT tasks, while the optimal retention rate remains modality-dependent.

\paragraph{Visualization.}
Figure~\ref{fig:example} presents qualitative comparisons. In the CT report case, MedARC preserves bilateral pleural effusion and adjacent lower-lobe atelectatic changes, whereas other methods omit, mislocalize, or hallucinate findings. In the MRI example, it uniquely recovers bilateral frontal white-matter abnormalities and mild paranasal-sinus mucosal thickening instead of producing generic normal descriptions. In VQA, MedARC is the only method that correctly returns \emph{no findings} for the negative CT case and reproduces the reference measurement of \(4\,\mathrm{mm}\), while the other methods predict unsupported abnormalities or incorrect sizes. These examples indicate that the fused saliency signals preserve clinically specific evidence during compression, while contextual merging retains complementary information and reduces plausible but unsupported predictions.

\subsection{Ablation Study}

\paragraph{Balance between Report Generation and VQA.}
Table~\ref{tab:ablation} evaluates the three importance signals by progressively adding visual--text similarity and VFM saliency to visual attention. Visual attention alone provides a strong baseline. Adding text similarity clearly improves VQA, but yields only limited or slightly negative changes in report generation, indicating that query conditioning favors targeted evidence over comprehensive clinical findings. VFM saliency compensates for this trade-off by preserving structurally distinctive regions that may not be strongly aligned with the text. The full model achieves the strongest overall report generation performance while maintaining the VQA gain

\paragraph{Robustness across Compression Levels.}
Figure~\ref{fig:ablation} shows that the full model remains stable under stronger compression on CT-RATE and is the only variant outperforming the uncompressed baseline across all three tasks near the default retention rate. MR-RATE degrades more noticeably under aggressive compression, suggesting greater sensitivity to token reduction in MRI. Overall, the three signals are complementary: text similarity improves query-specific reasoning, while VFM saliency preserves broader structural evidence.

\section{Conclusion \& Discussion}

We presented MedARC, a training-free framework for visual-token compression in 3D medical VLMs. By combining visual attention, visual--text similarity, and structural saliency from a frozen 3D VFM, MedARC adaptively retains important tokens and merges redundant ones. 
On CT-RATE and MR-RATE, it preserved or improved most metrics at about \(80\%\) token retention and remained competitive on CT tasks at approximately \(50\%\). 
Ablation results confirm that the three signals are complementary, while performance under aggressive MR compression suggests that the optimal token budget depends on the dataset and modality. Future work will extend MedARC to broader models, institutions, modalities, and rare findings, with further exploration of risk-aware allocation, robust saliency, and spatially constrained merging.

\bibliography{aaai2027}


\end{document}